\documentclass{article}
\usepackage{spconf,amsmath,amssymb,graphicx,hyperref,booktabs,makecell,caption,subcaption,tabularx,array,xcolor,multirow,eso-pic}

\definecolor{myblue}{HTML}{6C8EBF}
\definecolor{mygreen}{HTML}{82B366}
\definecolor{myorange}{HTML}{D79B00}
\definecolor{myyellow}{HTML}{D6B656}
\definecolor{myred}{HTML}{B85450}
\definecolor{mypurple}{HTML}{9673A6}

\title{Weather-R1: Logically Consistent Reinforcement Fine-Tuning for Multimodal Reasoning in Meteorology}
\name{
    Kaiyu Wu$^{1}$ \quad
    Pucheng Han$^{2}$ \quad
    Hualong Zhang$^{2}$ \quad
    Naigeng Wu$^{2}$ \quad
    Keze Wang$^{1}$\sthanks{Corresponding Author (e-mail: kezewang@gmail.com).}
}
  
\address{
    $^{1}$Sun Yat-sen University \\
    $^{2}$Guangdong Meteorological Observatory
}

\newcommand{\ieeenoticefirstpage}{%
    \AddToShipoutPicture*{%
        \AtPageUpperLeft{%
            \raisebox{-10mm}{%
                \makebox[0pt][l]{%
                    \hspace*{\dimexpr1in+\oddsidemargin\relax}%
                    \parbox{\textwidth}{\footnotesize
                    \copyright\ 2026 IEEE. Personal use of this material is permitted.
                    Permission from IEEE must be obtained for all other uses, in any current or
                    future media, including reprinting/republishing this material for advertising
                    or promotional purposes, creating new collective works, for resale or
                    redistribution to servers or lists, or reuse of any copyrighted component of
                    this work in other works.}}}}}}

\begin{document}
\ieeenoticefirstpage
\ninept
\maketitle
\begin{abstract}

While Vision Language Models (VLMs) show advancing reasoning capabilities, their application in meteorology is constrained by a domain gap and a reasoning faithfulness gap.
Specifically, mainstream Reinforcement Fine-Tuning (RFT) can induce Self-Contradictory Reasoning (Self-Contra), where the model's reasoning contradicts its final answer, which is unacceptable in such a high-stakes domain.
To address these challenges, we construct WeatherQA, a novel multimodal reasoning benchmark in meteorology.
We also propose Logically Consistent Reinforcement Fine-Tuning (LoCo-RFT), which resolves Self-Contra by introducing a logical consistency reward.
Furthermore, we introduce Weather-R1, the first reasoning VLM with logical faithfulness in meteorology, to the best of our knowledge.
Experiments demonstrate that Weather-R1 improves performance on WeatherQA by 9.8 percentage points over the baseline, outperforming Supervised Fine-Tuning and RFT, and even surpassing the original Qwen2.5-VL-32B.
These results highlight the effectiveness of our LoCo-RFT and the superiority of Weather-R1.
Our benchmark and code are available at \url{https://github.com/Marcowky/Weather-R1}.

\end{abstract}
%

\section{Introduction}

Amid escalating global climate change, weather forecasters must interpret extensive meteorological images and charts, and deliver reliable information~\cite{vicedo2021burden,clarke2022extreme}.
Although deep learning has advanced data-driven weather forecasting~\cite{chen2023fengwu,verma2024climode}, open-ended interpretation and reasoning still rely heavily on human experts.
Meanwhile, Vision Language Models (VLMs) have improved substantially in reasoning~\cite{hong2025glm,ma2025rlallvisualtriple,liu2025visualrftvisualreinforcementfinetuning,team2025kimi,tan2025reasonrftreinforcementfinetuningvisual}, opening up new possibilities for aiding forecasters with complex multimodal reasoning tasks in meteorology.
However, applying the general reasoning capabilities to meteorology, which demands specialization and reliability~\cite{vicedo2021burden,clarke2022extreme}, requires overcoming two challenges: the domain gap and the reasoning faithfulness gap.

To bridge the domain gap, as shown by prior works in fields such as medicine~\cite{li2024llava,pan2025medvlm,lai2025med} and mathematics~\cite{gao2025gllavasolvinggeometricproblem,li2025visionmatterssimplevisual,shi2024math}, it is essential to construct high-quality domain-specific instruction datasets for pre-training or fine-tuning~\cite{li2024llava,gao2025gllavasolvinggeometricproblem,li2025visionmatterssimplevisual,shi2024math}.
However, there is a shortage of high-quality datasets and benchmarks specifically for the meteorological domain.
In an effort to bridge this gap, we construct WeatherQA, a novel multimodal reasoning benchmark focused on meteorology, establishing a solid data foundation.

In terms of improving reasoning capabilities, DeepSeek-R1~\cite{guo2025deepseek} has demonstrated the effectiveness of Reinforcement Fine-Tuning (RFT)~\cite{liu2025visualrftvisualreinforcementfinetuning,tan2025reasonrftreinforcementfinetuningvisual}, which uses the direct evaluation of the final answer as a reward signal for reinforcement learning.
Similarly, MedVLM-R1~\cite{pan2025medvlm} and Med-R1~\cite{lai2025med} improved VLMs' performance on various medical multimodal tasks by applying RFT.
However, existing research has largely focused solely on optimizing the ``correctness of the final answer''~\cite{liu2025visualrftvisualreinforcementfinetuning,pan2025medvlm,lai2025med,yu2025dapo}, neglecting the quality of the reasoning process.
Our research reveals that this singular optimization paradigm causes models to exhibit a Self-Contradictory Reasoning (Self-Contra) phenomenon, where the reasoning process contradicts the final answer.
This unfaithful reasoning severely undermines the model's interpretability and trustworthiness, and is unacceptable in high-risk, high-precision meteorological applications~\cite{vicedo2021burden,clarke2022extreme}.

To address these challenges, we design the Logically Consistent Reinforcement Fine-Tuning (LoCo-RFT) paradigm to correct the inherent flaws of RFT.
This paradigm innovatively introduces a logical consistency reward to incentivize the model to generate reasoning paths that are logically consistent with the final answer.
Based on this paradigm and the WeatherQA dataset, we train our Weather-R1, the first VLM specifically tailored for multimodal reasoning tasks in meteorology, to the best of our knowledge.
Experimental results show that our Weather-R1 with 7B parameters achieves an accuracy of 52.9\% on the WeatherQA test set, a gain of 9.8 percentage points over the baseline model Qwen2.5-VL-7B~\cite{bai2025qwen2}.
Its performance not only surpasses that of Supervised Fine-Tuning (SFT) and RFT but also exceeds the original Qwen2.5-VL-32B.
Furthermore, our Weather-R1's generalization performance on the out-of-domain (OOD) ScienceQA benchmark~\cite{lu2022learn} shows a 4.98 percentage point improvement over the baseline.
These results validate the utility of our LoCo-RFT in enhancing multimodal reasoning performance.

The \textbf{main contributions} of this work are listed as follows:
\textbf{(i)} To the best of our knowledge, our Weather-R1 is the first logically consistent reasoning VLM designed specifically for meteorology, providing highly trustworthy and interpretable support for the field;
\textbf{(ii)} Our proposed WeatherQA benchmark is dedicated to multimodal reasoning in meteorology, bridging the gap of high-quality, multimodal data;
\textbf{(iii)} We introduce a novel paradigm, i.e., LoCo-RFT, to effectively suppress the Self-Contra in RFT and provide a new pathway for training more reliable reasoning models.

\section{Methodology}

\subsection{WeatherQA Benchmark}
\label{ssec:weatherqa_benchmark}

\begin{figure}[t]
    \centering
    \includegraphics[width=\linewidth]{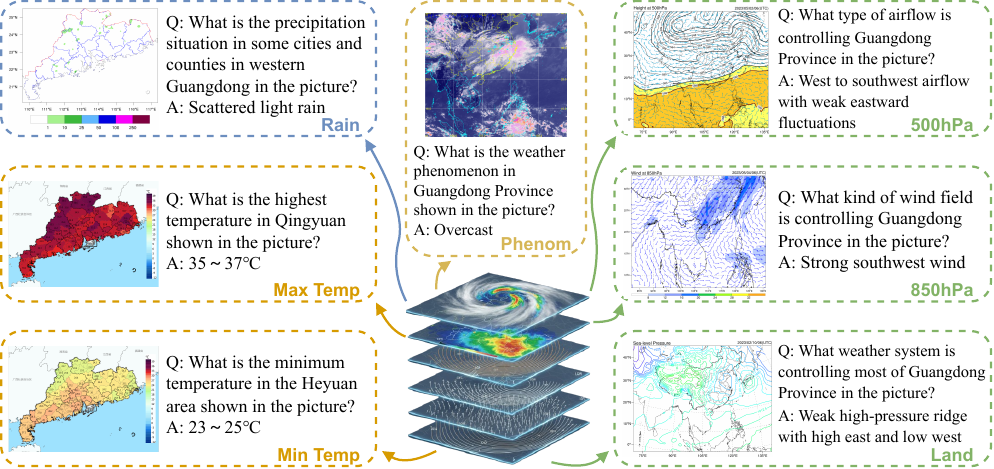}
    \vspace{-15pt}
    \caption{Data sample of our WeatherQA.
            The seven imaging modalities are: 
            24-hour cumulative precipitation map (\textcolor{myblue}{Rain}), 
            FY-2G satellite infrared cloud image (\textcolor{myyellow}{Phenom}), 
            daily maximum temperature map (\textcolor{myorange}{Max Temp}), 
            daily minimum temperature map (\textcolor{myorange}{Min Temp}), 
            500hPa geopotential height and wind field map (\textcolor{mygreen}{500hPa}), 
            850hPa wind field map (\textcolor{mygreen}{850hPa}), 
            and sea level pressure map (\textcolor{mygreen}{Land}). 
            These correspond to four themes: 
            \textcolor{myblue}{precipitation},
            \textcolor{myyellow}{weather phenomena},
            \textcolor{myorange}{temperature},
            and \textcolor{mygreen}{weather systems}, respectively.
            }
    \label{fig:weatherqa_example}
    \vspace{-10pt}
\end{figure}

The construction of our WeatherQA benchmark comprises 4 stages:

\noindent\textbf{Theme and Task Definition.}
In collaboration with meteorological experts, we define four themes for the benchmark:
precipitation, weather phenomena, temperature, and weather systems.
These themes correspond to seven specific imaging modality tasks (see Figure~\ref{fig:weatherqa_example}):
Rain: to identify the precipitation intensity;
Phenom: to identify the weather phenomena;
Max Temp \& Min Temp: to identify the maximum or the minimum temperature;
500hPa \& 850hPa \& Land: to identify weather systems at different pressure levels.
This design ensures the benchmark's domain coverage and task diversity.

\noindent\textbf{Data Preprocessing.}
Our raw data is sourced from weather analysis products.
After pairing images with their corresponding texts, we utilize DeepSeek-V3~\cite{liu2024deepseek} for theme segmentation in the text, and then extract pairs of \textlangle region, meteorological element\textrangle.

\noindent\textbf{Instruction Design and Generation.}
As in \cite{pan2025medvlm,lai2025med}, we design all instructions in a multiple-choice (single answer) format for automated quantitative evaluation.
First, for each \textlangle region, meteorological element\textrangle\, pair, we construct a question and its correct option using predefined task-specific templates.
Subsequently, we prompt GPT-4o~\cite{hurst2024gpt}, widely adopted in related works~\cite{openai2025gptoss120bgptoss20bmodel}, to generate the remaining distractor options based on the input image, question, and answer.
Note that a random 5\% sample of the dataset is validated by two meteorological experts, with both approval rates over 95\%, which justifies the technical rigor of our WeatherQA.

\noindent\textbf{Evaluation Protocol.}
We partition the dataset chronologically (train: 2017-2021, validation: 2022, test: 2023), yielding a 9:1:1 split.
To achieve a fair and comprehensive evaluation on our WeatherQA benchmark, we define a cross-task setting following \cite{pan2025medvlm,lai2025med}.
Specifically, the selected model is trained on a single task and evaluated on all seven imaging modality tasks.
The final accuracy for a specific task is then measured by the average accuracy of all models trained on each of the single tasks.

Ultimately, we construct WeatherQA, a multimodal multiple-choice benchmark for meteorology, comprising 15,400 entries that cover four themes and seven imaging modality tasks (see Figure~\ref{fig:weatherqa_example}).

\subsection{Reinforcement Fine-Tuning}
\label{ssec:selfcontra_and_rft}

Recent studies have confirmed that RFT can enhance the reasoning capabilities of large models in domains like mathematics~\cite{liu2025visualrftvisualreinforcementfinetuning,pan2025medvlm,lai2025med,shao2024deepseekmathpushinglimitsmathematical},
by functioning as an objective and concise reinforcement learning paradigm where the model freely explores the reasoning space and then uses the direct evaluation of the final answer as a reward signal for optimization.
It defines only two rewards: a format reward ($R_{Format}$), which requires the model to enclose the reasoning process ($rp$) in ``think'' tags, and the final answer ($fa$) in ``answer'' tags;
and an accuracy reward ($R_{Acc}$), which evaluates whether the $fa$ is correct.
For each reward, a value of 1 is assigned for compliance and 0 otherwise.
This setup ensures that the model maintains accuracy while adhering to a structured output format during the reasoning process.
RFT is often optimized using the Group Relative Policy Optimization (GRPO) algorithm~\cite{shao2024deepseekmathpushinglimitsmathematical}, which does not require a separate critic model to estimate state-value functions.
Instead, it provides efficient supervisory signals by directly comparing groups of responses.
Specifically, GRPO generates $G$ responses, each formulated as a pair \textlangle $rp$, $fa$\textrangle, for a given question $q$ based on the current policy, which are assigned rewards $\{r_1, r_2, ..., r_G\}$.
Then, for the i-th response, its corresponding advantage value $A_i$ is obtained by normalizing the relative rewards within the group:

\vspace{-11pt}
\begin{align}
    A_{i}=\frac{r_i-mean(\{r_1, r_2, ..., r_G\})}{std(\{r_1, r_2, ..., r_G\})}
\end{align}
\vspace{-6pt}

\noindent\textbf{Self-Contradictory Reasoning in RFT.}
Previous work~\cite{liu2024selfcontradictoryreasoningevaluationdetection,mündler2024selfcontradictoryhallucinationslargelanguage} noted that large models can exhibit the Self-Contradictory Reasoning (Self-Contra) phenomenon during reasoning.
When applying RFT to multimodal reasoning in meteorology, we find this phenomenon to be particularly prominent:
In the multiple-choice scenarios of WeatherQA, we observe numerous contradictory cases where the $rp$ logically points to option C, but the $fa$ is option A (see Figure~\ref{fig:case_study}).
We categorize this phenomenon into three main types:
$Type~1$: Correct $rp$ leads to an incorrect $fa$;
$Type~2$: Incorrect $rp$ leads to a correct $fa$; and
$Type~3$: Incorrect $rp$ leads to a different incorrect $fa$ or the $rp$ itself is inconclusive.

\begin{table}[t]
\caption{Prompt used for judge model to obtain $fa_{rp}$. The placeholders \{$Question$\}, \{$Choices$\}, and \{$rp$\} are replaced with the actual Question, Choices, and $rp$, respectively.}
\label{tab:locoprompt}
\vspace{-7pt}
\centering
\footnotesize
\begin{tabular}{p{0.95\linewidth}}
\toprule
Your task is to select the option best supported by the given reasoning process. \\
Directly output the uppercase letter of the selected option. If the reasoning process does not correspond to any of the options, output ``Cannot be determined''. \\
{[Input]}: Question: \{$Question$\}\textbackslash n Choices: \{$Choices$\}\textbackslash n Reasoning process: \{$rp$\}\\
{[Output]}: \\
\bottomrule
\end{tabular}
\vspace{-5pt}
\end{table}

\begin{figure}[tb]
    \centering
    \begin{minipage}[t]{0.45\linewidth}
        \vspace{2pt} 
        \centering
        \setlength{\tabcolsep}{1.5pt}
        \renewcommand{\arraystretch}{1.1} 
        \begin{tabular}{lcc}
        \hline
        Type     &   \makecell{GRPO \\ Avg } &   \makecell{DAPO \\ Avg } \\ 
        \hline
        Self-Contra & 33.23  & 29.93  \\
        $Type~1$      & 8.04   & 10.36  \\
        $Type~2$      & 10.90  & 8.64  \\
        $Type~3$      & 14.30   & 10.93   \\ 
        \hline
        \end{tabular}
        \vspace{-2pt} 
        \subcaption{}\label{fig:self_contra}
    \end{minipage}
    \hfill
    \begin{minipage}[t]{0.52\linewidth}
        \vspace{0pt}
        \centering
        \includegraphics[width=\linewidth]{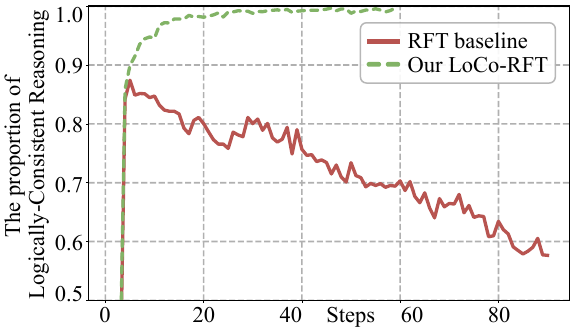}
        \vspace{-15pt} 
        \subcaption{}\label{fig:loco_change}
    \end{minipage}
    \vspace{-6pt}
    \caption{\textbf{(a)} Self-Contra statistics. Each column shows the average proportion (\%) of Self-Contra responses per algorithm across seven tasks. \textbf{(b)} Proportions of logically consistent reasoning responses during RFT and our LoCo-RFT training. In early training, unstable response formats prevent $rp$ extraction, which are not counted as consistent, leading to the lower proportion.}
    \label{fig:overall}
    \vspace{-10pt}
\end{figure}

To quantify this phenomenon, we use the powerful open-source large model gpt-oss-20b~\cite{openai2025gptoss120bgptoss20bmodel} as a judge (see Table~\ref{tab:locoprompt} for prompt).
Given an input of \textlangle Question, Choices, $rp$\textrangle, it is tasked with selecting the option best supported by the reasoning, referred to as $fa_{rp}$.
If no option can be selected, it outputs ``Cannot be determined.''
Subsequently, we compare the $fa_{rp}$ with the model's $fa$; a mismatch is classified as a Self-Contra.
As shown in Figure~\ref{fig:self_contra}, we measure the incidence of this phenomenon after RFT training and find that the Self-Contra proportion is approximately 30\% across all seven tasks in WeatherQA, regardless of whether the GRPO or DAPO~\cite{yu2025dapo} algorithm is used.
Furthermore, as depicted in Figure~\ref{fig:loco_change}, the proportion of logically consistent responses gradually decreases as RFT training progresses, indicating that Self-Contra becomes more frequent.

We argue that the cause of Self-Contra in RFT lies in the singularity of its optimization objective.
Besides the format reward, the only optimization signal comes from the correctness of the final answer.
This may reward a self-contradictory response that happens to be correct while penalizing a logically consistent response that is incorrect due to the model's capability limitations.
This optimization objective conflicts with the logical consistency learned during the pre-training phase, causing the model to learn how to ``guess'' the right answer through unfaithful reasoning patterns to gain rewards.

\subsection{Logically Consistent Reinforcement Fine-Tuning}

\begin{figure}[t]
    \centering
    \includegraphics[width=\linewidth]{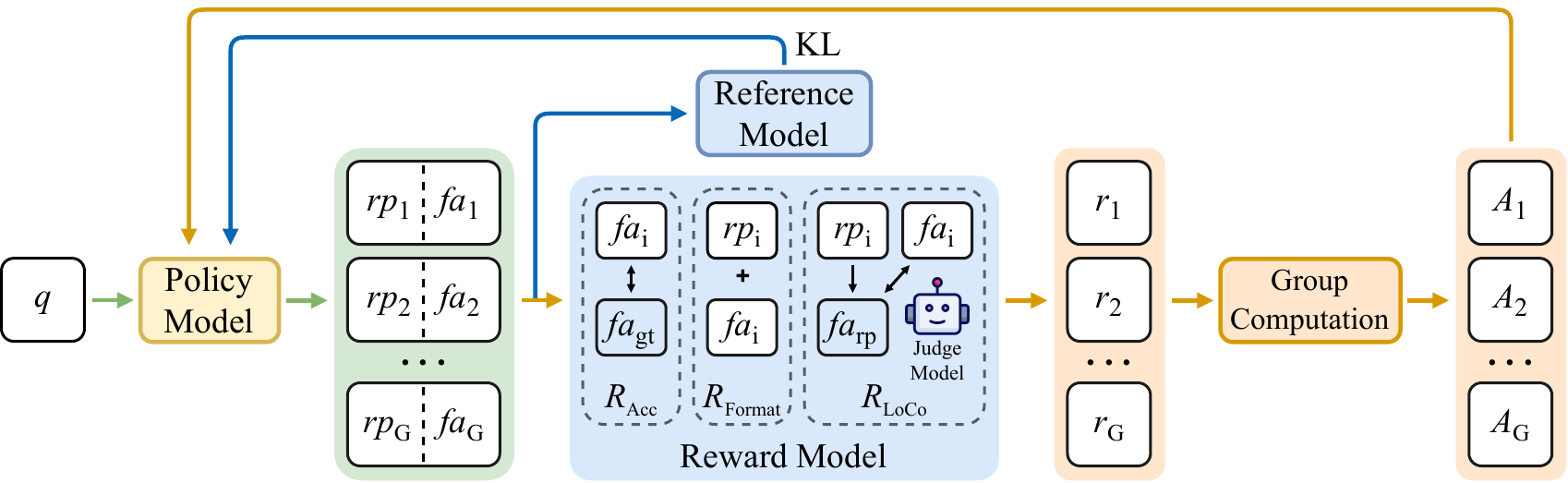}
    \vspace{-15pt}
    \caption{The LoCo-RFT paradigm. We introduce an additional LLM-assisted logical consistency reward, $R_{LoCo}$, to suppress the Self-Contra phenomenon.}
    \label{fig:locorft}
    \vspace{-10pt}
\end{figure}

To address the Self-Contra in RFT, we propose Logically Consistent Reinforcement Fine-Tuning (LoCo-RFT) by introducing a new reward dimension, i.e., the logical consistency reward ($R_{LoCo}$), to guide the model in maintaining the faithfulness of its reasoning process while pursuing answer correctness (see Figure~\ref{fig:locorft}).
The formula for $R_{LoCo}$ is as follows:

\vspace{-15pt}
\begin{align}
R_{LoCo} =
    \begin{cases}
    1, & \text{if } fa_{rp}=fa \text{ and } R_{Format}=1,\\
    0, & \text{otherwise}.
    \end{cases}
\end{align}
\vspace{-9pt}

\noindent where $fa_{rp}$, as defined in Section~\ref{ssec:selfcontra_and_rft}, is the option extracted by the judge model that is best supported by the $rp$.
This task-agnostic extraction (prompt in Table~\ref{tab:locoprompt}) leads to a low risk of bias.
The condition $R_{Format}=1$ ensures the judge model only receives the well-formatted $rp$ without $fa$, thereby preventing potential misjudgment.

We continue to use open-source gpt-oss-20b as the judge model.
For validation, a sample of 300 model responses, including instances from all seven tasks and all three types of Self-Contra, is randomly selected and manually annotated with their $fa_{rp}$.
Upon comparing these annotations with the $fa_{rp}$ from gpt-oss-20b, we find a Cohen's Kappa coefficient of 0.9778, which indicates almost perfect agreement and thus confirms the high reliability of gpt-oss-20b as a judge.

Finally, the total reward function for LoCo-RFT is a weighted sum of three components, i.e., format reward ($R_{Format}$), logical consistency reward ($R_{LoCo}$), and accuracy reward ($R_{Acc}$), with empirically corresponding weights of 0.1, 0.3, and 0.6.
We validate this weighting in our ablation study (Section~\ref{ssec:ablation_study}), which confirms its effectiveness in balancing answer accuracy with logical consistency.

\section{Experiments}


\noindent\textbf{Datasets \& Tasks.}
We use WeatherQA as our training dataset and evaluation benchmark, following its defined cross-task protocol (see Section~\ref{ssec:weatherqa_benchmark}).
Additionally, to measure the model's OOD generalization capability, we curate a test set from ScienceQA~\cite{lu2022learn}, which consists of 324 multiple-choice questions related to weather and climate.
Multiple-choice accuracy is used as the metric for all experiments.

\noindent\textbf{Implementation Details.}
We initialize our Weather-R1-7B with the weights of Qwen2.5-VL-7B-Instruct~\cite{bai2025qwen2} and perform full-parameter LoCo-RFT on 4×A100 GPUs.
For GRPO, we sample 5 responses per group with a temperature of 1.0.
The batch size is set to 4 per GPU, with 8 steps of gradient accumulation.
Then, we train the model for multiple epochs, evaluating its performance on the validation set after each epoch to select the best-performing checkpoint.
Besides, we deploy gpt-oss-20b~\cite{openai2025gptoss120bgptoss20bmodel} as the judge model using vLLM~\cite{kwon2023efficient} engine, which only requires about 20GB of VRAM, and set the concurrency for reward calculation to 256 to reduce the computational time overhead of $R_{LoCo}$.
Since the policy update constitutes the majority of the time cost in RFT, our average single-step training time increases by only 0.55\% (from 970.7s to 976.0s).
Inference-time requires no judge, incurring no additional cost.

\noindent\textbf{Baselines.}
For a comprehensive and fair comparison, we select two categories of baseline models:
(i) Zero-shot VLMs: This includes models from the LLaVA-1.6~\cite{liu2024llavanext} series and the Qwen2.5-VL~\cite{bai2025qwen2} series.
(ii) Fine-tuned VLMs: We train SFT-7B and RFT-7B models under the same settings to serve as comparative baselines.

\subsection{Main Result}

\begin{table}[t]
    \centering
    \caption{Performance on the WeatherQA test set.
    For the Fine-Tuned VLMs, we employ a cross-task setting of WeatherQA.
    }
    \vspace{-7pt}
    \footnotesize
    \setlength{\tabcolsep}{1.5pt} 
    \begin{tabular}{>{\raggedright\arraybackslash}p{2.1cm}|*{7}{>{\centering\arraybackslash}p{2.3em}}|>{\centering\arraybackslash}p{3em}}
        \hline
        Model\textbackslash Task & \makecell{500 \\ hPa} & \makecell{850 \\ hPa} & Land & Rain & \makecell{Phe- \\ nom} &
        \makecell{Min \\ Temp} &
        \makecell{Max \\ Temp} &
        Overall \\
        \hline
        \multicolumn{9}{c}{Zero-shot VLMs} \\
        \hline
        LLaVA-v1.6-7B   & 34.5 & 26.5 & 23.0 & 37.0 & 27.5 & 25.0 & 20.0 & 27.6 \\
        LLaVA-v1.6-13B  & 35.5 & 35.0 & 28.0 & 22.0 & 60.0 & 23.5 & 25.0 & 32.7 \\
        Qwen2.5-VL-3B   & 38.0 & 45.5 & 36.0 & 23.5 & 52.5 & 60.0 & 44.0 & 42.8 \\
        Qwen2.5-VL-7B   & 35.5 & 39.5 & 30.5 & 35.0 & 57.0 & 56.5 & 47.5 & 43.1 \\
        \hline
        \multicolumn{9}{c}{Fine-tuned VLMs} \\
        \hline
        SFT-7B          & 43.8 & 42.9 & \textbf{49.9} & 40.6 & 62.0 & 58.4 & 50.8 & 49.8 \\
        RFT-7B          & 43.1 & \textbf{48.4} & 47.9 & 41.5 & \textbf{64.6} & 62.9 & 51.4 & 51.4 \\
        Weather-R1-7B   & \textbf{44.8} & 47.5 & 46.6 & \textbf{41.9} & 64.4 & \textbf{67.7} & \textbf{57.4} & \textbf{52.9} \\ 
        \hline
    \end{tabular}
    \vspace{-5pt}
    \label{tab:main_result}
\end{table}

\begin{table}[t]
    \centering
    \caption{Generalization performance on ScienceQA.}
    \vspace{-7pt}
    \footnotesize
    \setlength{\tabcolsep}{3pt} 
    \begin{tabular}{>{\raggedright\arraybackslash}p{1.8cm}|
                    *{4}{>{\centering\arraybackslash}p{4em}}}
        \hline
        Metric\textbackslash Model & \makecell{Qwen2.5- \\ VL-7B} & \makecell{SFT-7B} & \makecell{RFT-7B} & \makecell{Weather- \\ R1-7B} \\
        \hline
        Accuracy & 81.48 & 78.22 & 79.54 & 86.46 \\
        \hline
    \end{tabular}
    \vspace{-10pt}
    \label{tab:scienceqa_result}
\end{table}

\noindent\textbf{Performance on WeatherQA.}
The cross-task performance of different models on WeatherQA is presented in Table~\ref{tab:main_result}.
The experimental results show that Weather-R1-7B achieves the best overall average accuracy of 52.9\%, an improvement of 9.8 percentage points over the baseline Qwen2.5-VL-7B (43.1\%).
Notably, with a 7B parameter scale, Weather-R1-7B surpasses the original 32B-parameter Qwen2.5-VL (52.0\%).
More importantly, compared to other Fine-Tuned VLMs, Weather-R1-7B's performance surpasses SFT-7B (49.8\%) and RFT-7B (51.4\%) by 3.1 and 1.5 percentage points, respectively.
On individual tasks, Weather-R1-7B performs best on 500hPa, Rain, Min Temp, and Max Temp tasks, and ranks second on the 850hPa and Phenom tasks.
These results demonstrate the superiority of our LoCo-RFT paradigm, highlighting its ability to foster strong cross-task capabilities and the critical role of logically consistent reasoning in meteorology.

\noindent\textbf{Generalization Performance on ScienceQA.}
The generalization performance of different models on the OOD benchmark ScienceQA is shown in Table~\ref{tab:scienceqa_result}.
Although Weather-R1-7B is not trained on its dataset, it still achieves an accuracy of 86.46\%, an improvement of 4.98 percentage points over the baseline Qwen2.5-VL-7B (81.48\%).
In stark contrast, compared to the baseline, the SFT-7B and RFT-7B show a decrease in accuracy of 3.26\% and 1.94\%, respectively.
This result indicates that Weather-R1-7B, trained with LoCo-RFT, not only enhances in-domain logically consistent reasoning capabilities but also successfully generalizes this ability to OOD tasks.

\subsection{Ablation Studies}
\label{ssec:ablation_study}
To demonstrate the contribution of our LoCo-RFT's key components, i.e., the judge model, reward weights, and GRPO sample number $G$, we have conducted the following ablation studies in Table~\ref{tab:ablation_study}.
Specifically, replacing our judge model with larger or smaller alternatives, Qwen3-32B and Qwen3-4B~\cite{qwen3technicalreport}, results in a slight performance decrease (Rows 2 \& 3) but still outperforms the RFT baseline (Row 6).
This result confirms the superiority of our selected judge, i.e., gpt-oss-20b, and demonstrates LoCo-RFT's robustness to the different judge models.
Besides, altering the balance to favor either $R_{Acc}$ or $R_{LoCo}$ leads to minor performance drops (Rows 4 \& 5), though both remain superior to standard RFT.
This indicates our chosen weights strike an effective balance between the two objectives, and that LoCo-RFT is also insensitive to the hyperparameters.
Finally, a smaller sample number $G$ degrades performance (Row 8), while a larger $G$ can bring some benefits (Row 7), but increases the computational cost.
This demonstrates that our LoCo-RFT scales effectively with the sample number, and our selected $G$ represents a reasonable trade-off between performance and efficiency.

\begin{table}[t]
    \centering
    \caption{Ablation study for our LoCo-RFT. All models are trained on the 500hPa task and evaluated on WeatherQA and ScienceQA.}
    \vspace{-7pt}
    \label{tab:ablation_study}
    \footnotesize
    \setlength{\tabcolsep}{3pt}
    \newcolumntype{C}[1]{>{\centering\arraybackslash}p{#1}}
    \begin{tabular}{l|C{2.8em}C{2.8em}|c|cc}
        \hline
        \multirow{2}{*}{\makecell{Judge \\ Model}} & \multicolumn{2}{c|}{Weight of} & \multirow{2}{*}{\makecell{Sample \\ num $G$}} & \multicolumn{2}{c}{Accuracy of} \\
        & $R_{LoCo}$ & $R_{Acc}$ & & WeatherQA & ScienceQA \\
        \hline
        \textbf{gpt-oss-20b} & \textbf{0.3} & \textbf{0.6} & \textbf{5} & 57.29 & 87.65 \\
        \underline{Qwen3-32B}   & 0.3 & 0.6 & 5 & 56.57 (-0.72) & 85.49 (-2.16) \\
        \underline{Qwen3-4B}   & 0.3 & 0.6 & 5 & 56.93 (-1.00) & 86.73 (-0.92) \\
        gpt-oss-20b & \underline{0.1} & \underline{0.8} & 5 & 56.07 (-1.22) & 87.00 (-0.65) \\
        gpt-oss-20b & \underline{0.5} & \underline{0.4} & 5 & 56.57 (-0.72) & 84.57 (-3.08) \\
        \underline{None (RFT)}  & \underline{0}   & \underline{0.9} & 5 & 54.64 (-2.65) & 80.86 (-6.79) \\
        gpt-oss-20b & 0.3 & 0.6 & \underline{8} & 58.43 (+1.14) & 89.20 (+1.55) \\
        gpt-oss-20b & 0.3 & 0.6 & \underline{3} & 56.14 (-1.15) & 87.04 (-0.61) \\
        \hline
    \end{tabular}
    \vspace{-5pt}
\end{table}

\subsection{Analysis of Cross-Task Results}

\begin{figure}[t]
    \vspace{-3pt}
    \centering
    \includegraphics[width=\linewidth]{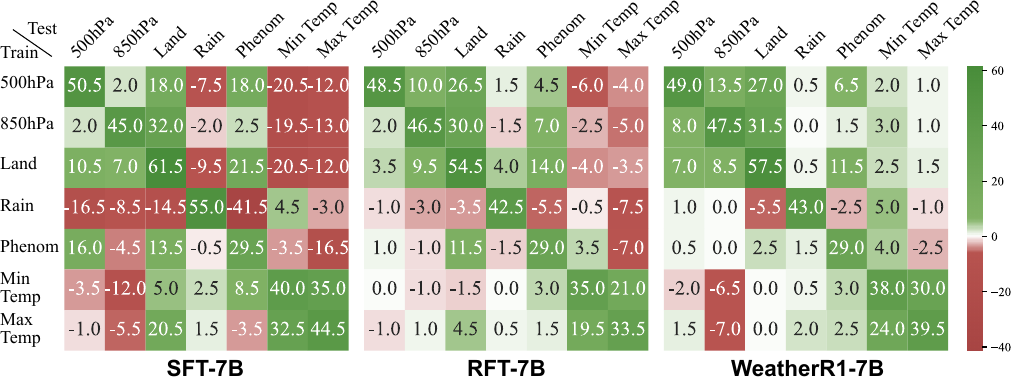}
    \vspace{-15pt}
    \caption{Detailed cross-task performance on WeatherQA. For each model, rows represent the training task, and columns represent the testing task. The value in each cell denotes the percentage change in performance compared to the Qwen2.5-VL-7B baseline. \textcolor{mygreen}{Green} signifies improvement, while \textcolor{myred}{red} signifies a decline.}
    \label{fig:result_detail}
    \vspace{-10pt}
\end{figure}

Figure~\ref{fig:result_detail} shows the detailed cross-task performance of different models on WeatherQA.
For in-task performance (the diagonal cells), SFT-7B exhibits more significant performance gains than the reinforcement learning-based models.
However, for cross-task performance (the off-diagonal cells), SFT-7B also leads to more frequent performance degradation, whereas Weather-R1-7B shows fewer instances of performance decline than RFT-7B.
We count the number of cells where performance degraded (marked in red) for each model.
The results show that SFT-7B and RFT-7B have 23 and 19 instances of performance degradation, respectively, while Weather-R1-7B has only 7.
This indicates that LoCo-RFT endows the model with stronger cross-task transferability, enabling it to better adapt to the diverse imaging modalities and specific tasks in meteorology.

\subsection{Evaluation for Logically Consistent Reasoning}

We evaluate logical consistency using the same setup in Section~\ref{ssec:selfcontra_and_rft}.
In Table~\ref{tab:self_contra_exp}, compared to RFT-7B, our Weather-R1-7B's Self-Contra proportion is lower by 31.41\% and 17.55\% on WeatherQA and ScienceQA.
Besides, in Figure~\ref{fig:loco_change}, the proportion of logically consistent responses gradually increases as our LoCo-RFT training progresses.

\begin{figure}[t]
    \centering
    \includegraphics[width=\linewidth]{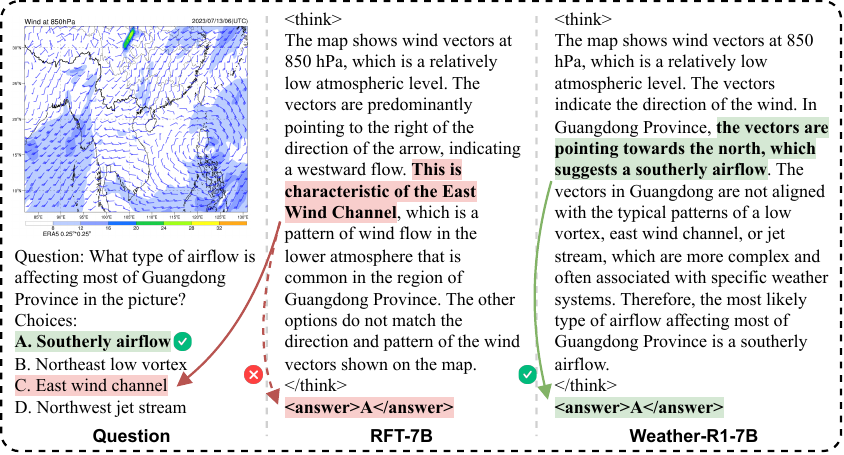}
    \vspace{-15pt}
    \caption{Response Comparison of RFT-7B and Weather-R1-7B.
    }
    \label{fig:case_study}
    \vspace{-10pt}
\end{figure}

A case study in Figure~\ref{fig:case_study} visually illustrates this difference.
For the same question, the RFT-7B's reasoning process leads to an incorrect option C, yet its final answer is option A, which contradicts the reasoning, revealing a clear Self-Contra.
In contrast, Weather-R1-7B generates a logically coherent chain of thought:
It analyzes the direction of the wind vectors, identifies a ``southerly airflow,'' discusses the errors in other options, and ultimately provides a correct answer that is consistent with its reasoning.
This clearly demonstrates that LoCo-RFT can effectively guide the model to maintain logical consistency in its reasoning during training, thereby suppressing the occurrence of the Self-Contra phenomenon.

\begin{table}[t]
    \centering
    \caption{Self-Contra statistics on WeatherQA and ScienceQA.}
    \vspace{-7pt}
    \footnotesize
    \setlength{\tabcolsep}{2.5pt} 
    \begin{tabular}{l|cc|cc}
        \hline
        Model & \multicolumn{2}{c|}{RFT-7B} & \multicolumn{2}{c}{WeatherR1-7B} \\ 
        \hline
        Type\textbackslash Benchmark & \makecell{WeatherQA} & \makecell{ScienceQA} & \makecell{WeatherQA} & \makecell{ScienceQA} \\ 
        \hline
        Self-Contra   & 33.23 & 20.02 & 1.82 & 2.47 \\
        $Type~1$        & 8.04  & 9.08  & 0.29 & 0.66 \\
        $Type~2$        & 10.90 & 4.94  & 0.26 & 0.97 \\
        $Type~3$        & 14.30  & 6.00  & 1.28 & 0.84 \\
        \hline
    \end{tabular}
    \label{tab:self_contra_exp}
    \vspace{-5pt}
\end{table}

\section{Conclusion}
\label{sec:conclusion}

In this work, we introduce WeatherQA, a novel multimodal reasoning benchmark for meteorology.
Furthermore, we identify the Self-Contra issue in RFT and propose a novel LoCo-RFT paradigm to mitigate it by rewarding faithful reasoning.
Our Weather-R1 demonstrates the effectiveness of this paradigm by significantly reducing Self-Contra proportion while achieving state-of-the-art performance on the WeatherQA benchmark.
This research provides a new paradigm for developing more reliable and interpretable VLMs for high-stakes specialized fields.
For future work, we plan to extend the LoCo-RFT paradigm to open-ended generation tasks.
The ability to generate reliable, free-form text is crucial across many specialized fields, not just meteorology.
Ensuring logical faithfulness in these scenarios thus validates our approach as a generalizable paradigm for creating trustworthy AI assistants for high-stakes domains.

\vfill\pagebreak

\bibliographystyle{IEEEbib}
\bibliography{strings,refs}

\end{document}